\documentclass[letterpaper, 10 pt, conference]{ieeeconf}  

\IEEEoverridecommandlockouts                              

\usepackage{ftnxtra}
\usepackage{graphicx}
\usepackage{amsmath}
\usepackage{amssymb}
\usepackage[hidelinks]{hyperref}
\usepackage{float}

\title{\LARGE \bf
Dynamic and Static Object Detection Considering Fusion Regions and Point-wise Features
}

\author{Andr\'{e}s E. G\'omez H.$^{1}$, Thomas Genevois$^{1}$, Jerome Lussereau$^{1}$ and Christian Laugier$^{1}$
\thanks{*This work was supported by the FUI STAR project}
\thanks{\small
$^{1}$ Authors are with Univ. Grenoble Alpes, INRIA Grenoble Rh{\^o}ne-Alpes, Chroma Team, 38334 Montbonnot, France. Correspondence: {\tt\small andres.gomez-hernandez@inria.fr, aegh@ieee.org}}%
}

\begin{document}

\maketitle
\thispagestyle{empty}
\pagestyle{empty}


\begin{abstract}

Object detection is a critical problem for the safe interaction between autonomous vehicles and road users. Deep-learning methodologies allowed the development of object detection approaches with better performance. However, there is still the challenge to obtain more characteristics from the objects detected in real-time. The main reason is that more information from the environment's objects can improve the autonomous vehicle capacity to face different urban situations. This paper proposes a new approach to detect static and dynamic objects in front of an autonomous vehicle. Our approach can also get other characteristics from the objects detected, like their position, velocity, and heading. We develop our proposal fusing results of the environment's interpretations achieved of YoloV3 and a Bayesian filter. To demonstrate our proposal's performance, we asses it through a benchmark dataset and real-world data obtained from an autonomous platform. We compared the results achieved with another approach.

\end{abstract}


\section{INTRODUCTION}

Autonomous Vehicles (\textit{AV}) will only become a success if people accept them. However, as Thomas et al. concluded in \cite{thomas2020}, some persons still do not believe in Av's performance. In fact, behind the AV development still exists several technical challenges such as software complexity, real-time data analytics, testing, and verification \cite{hussain2018}. Nonetheless, a first step to win the users' confidence can be to define new models for the AV's perception-system that improve its environment understand.

Different mathematical models integrate the perception system. Some of them permit the AV to find, classify and track several objects. The models decode sensor data collected from the physical world based on some requirements. Therefore, the sensors are a central part of the perception system. However, according to the sensor type, the model outcome could present some troubles. For example, an RGB camera sensor generally is a common choice given its cost and the significant amount of information available inside its field of view. Nevertheless, This sensor is vulnerable to light changes and weather conditions. Consequently, the models have to consider fusing data from different sensors to obtain a result free of a single sensor's limitations.

Critical models for the safety around an AV can improve their results by fusing information from different sensor sources. Object detection models are an example of it. By default, the development of some object detection models considers only the use of RGB cameras. In that models, the image detections are limited only to located and classified the objects inside a camera field of view. However, Lidar measures from the environment could help obtain more information concerning the same object detected by the RGB cameras (e.g., object position 2D). 

We propose a new approach to detect dynamic and static objects in an urban environment. Our approach also gets other characteristics from the objects detected, such as position, velocity, and heading. We found the objects' information fusing the environment's interpretations from two different approaches. The first approach is the YoloV3 object detector \cite{Redmon2018}. YoloV3 obtains a likely classification and position from each object inside an image achieved with an RGB camera. The second approach is the Bayesian filter known as the Conditional Monte Carlo Dense Occupancy Tracker(\textit{CMCDOT})\cite{rummelhard2015}. This approach achieves dynamic and static information from the urban environment through a lidar sensor. Both approaches work in real-time. Our main contributions can be summarized as follows:

\begin{itemize}
    \item Fusion of the results achieved from an object detector and a Bayesian filter to find dynamic and static objects in an urban environment.
    \item To get more characteristics about the objects detected by processing point-wise features inside fusion regions.
\end{itemize}

The paper organization has the following structure: Section II reviews the related work. Section III details the definition of the fusion regions and the point-wise feature processing. Section IV shows the experimental results. Finally, section V provides concluding remarks.


\section{RELATED WORK}

We review related works with a focus on object detections for AV. Specifically, we considered the works with two essential aspects: i) the number of sensors employed, ii) and the deep learning approaches examined to object detection. In the first aspect, we found the work presented by Liang et al. in \cite{liang2020}. They considered that the range image representation is a feasible alternative to object detection. Therefore, they developed a 3D object detector framework called RangeRCNN, based on range images. However, despite good performance obtained by RangeRCNN in the KITTI benchmark, it only contemplates lidar data as input.

Arnold et al. in \cite{arnold2019} reviewed several 3D object detection approaches on AV. Moreover, the authors also studied different sensor technologies and standards datasets. They consider that 2D detections on the image plane are not enough for a safety-critical system as an AV. Therefore, The authors proved that an object detection approach needs to fuse information from different sensors with models executed in real-time. This fact is the basis of our proposal.

The sensors' information and the fusion models applied in them are essential to determine the AV's safety. In \cite{wang2019}, Wang, Wu, and Niu analyzed typical sensors and several multi-sensor fusion strategies used in the last years on AVs. The authors confirmed the need to fuse sensors' information because it is necessary to avoid a single sensor's limitations and uncertainties. This conclusion is a found fact in several papers consulted.

Rangesh and Trivedi in \cite{rangesh2019} present the $M^{3}OT$ framework. This framework can accept an object proposal from different sensor modalities to track it. To obtain the object track recommendations, Rangesh and Trivedi considered a tracking-by-detection approach. This approach uses a nearly-fusion model to associate the objects detected by each sensor. The result of the nearly-fusion model is a merged representation from the raw sensor data. 

Zhu et al. in \cite{zhu2017} presented a literature review about several techniques used to environment perception for intelligent vehicles. The document summarizes different methods with their pros and cons. According to this, the deep-learning methods adopted for the scene understanding obtained a relevant validation. However, the authors considered that these deep-learning methods should consider motion and depth information to improve its accuracy. Consequently, an occupancy grid approach could be helpful to complement these deep-learning methods.

For the second aspect, it was possible to find different deep learning approaches used on object detection problems. For example, Pang and Cao in \cite{pang2019} analyzed some object detection methods based on deep learning. They compare typical CNN-based architectures focused on a specific use case, pedestrian detection. One conclusion of this work is that object detection using deep-learning approaches has several challenges. The occlusion, scale variation, and deformation in the objects are some of them.

In \cite{zhao2019}, Zhao et al. provided the result of a systematic review, which analyzed object detection using deep-learning models on different applications. They compared each model through experiments. According to the experiments' results, the authors also concluded two main challenges on object detection. The first one is real-time object detection, and the second one is to extend the classical methods for 2D to 3D object detection.   

Additionally, we did find some proposals where the authors considered deep-learning approaches for the fusion of sensors' information. For example, Nobis et al. in \cite{nobis2019} proposed the fusion between radar and camera sensor data using a neural network approach. Their proposal, knowledged as Camera Radar Fusion Net (CRF-Net), merges a 2D object detector with projected sparse radar data in the image plane. Moreover, CRF-Net filters the data obtained in the image plane to improve the object detection performance. However, CRF-Net does not work in real-time.

In an AV, the perception system not only needs to detect the objects around the urban environment. This system need also consider if the object is moving or not. Given this fact, Siam et al. in \cite{siam2018} proposed a new model that combines motion and appearance cues using a single convolutional network. Therefore, the outcome obtained with the Moving Object Detection Network (MODNet) is the static/moving classification for objects. Moreover, another contribution of this work is the KITTI MOD dataset. This dataset acts as a benchmark on motion detection on the KITTI dataset.

Finally, Liang et al. in \cite{liang2019} exploited multiple related tasks for 3D object detection using multiple sensors. Moreover, They proposed to predict dense depth from multiple sensor data and use this result to find near correspondences between multiple sensor feature maps. These predictions are possible through a network architecture that fuses the detections from a LiDAR point cloud and an RGB image. This network architecture considers 2D and 3D object detection, ground estimation, and depth completion.

After all the works analyzed previously, it is clear that object detection on AV is not a trivial problem. Therefore, it is needed to obtain and process enough scene information from diverse sensor sources. Moreover, the fusion models used to merge all the information obtained must have a real-time performance. Consequently, combining distinct methodologies like deep-learning and Bayesian filters, it is possible to complete these requirements. Our paper proposes a fusion model that combines the YoloV3 object detector with an occupancy map obtained from the CMCDOT approach. The following section explains in detail the methodology used for our proposal.


\section{METHOD}

\begin{figure*}[h!]
\centering
\framebox{\includegraphics[height=8.5cm, width=\textwidth]{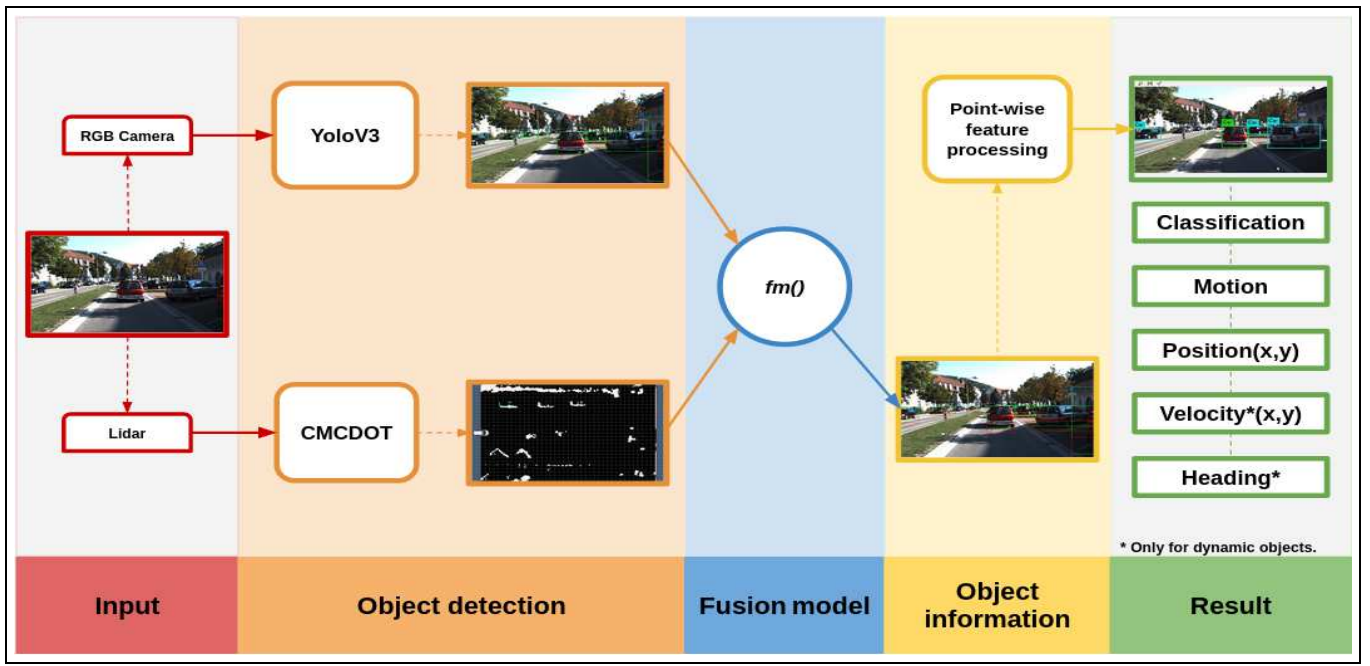}}
\caption{The figure shows the methodology's block diagram used in our proposal. The \textit{input part} obtains raw data from an urban scene using two sensors: lidar and an RGB camera. Eventually, in the \textit{object detection part}, we utilized two approaches to look for the objects in front of the Ego-car: YoloV3 and CMCDOT. Then, we merge the result take from YoloV3 and CMCDOT approaches in the \textit{Fusion model part}. Afterward, we processed the previous area's outcome in the \textit{object information part} to found the classification, motion estimation, position, velocity, and heading from the objects\footnotemark. Finally, we present all the information found in the \textit{result part}.}
\label{fig:me}
\end{figure*}

Based on the evidence found in the related works, we define the methodology used in our proposal. It is composed of five parts described in the block diagram of figure \ref{fig:me}. In this figure, it is possible to observe an ordered sequence of subprocesses connected between each section of the block diagram. In the remainder of this section, we will explain each subprocess and its implications in the methodology.

\subsection{Input}

An object detection approach needs to obtain information from distinct sources to warranty an adequate performance over different situations. Considering this fact, we decided to use an RGB camera and a lidar sensor to object detection. These two kinds of sensors have been used in several works from state of the art, mainly because each one supports the other in its functional disadvantages. One instance could be RGB cameras' problem with lack of depth information or lidar's issue with the loss of object details in the point cloud. Moreover, raw data information about these two sensors is available in diverse datasets like in \cite{Geiger2013} or \cite{caesar2020}. 

\subsection{Object detection}

We used the two sensors' information as inputs for the YoloV3 and the CMCDOT approaches. YoloV3 is an object detector that takes the RGB camera images to classify and locate the objects presented inside the image considering a probability.  The main reason to have chosen this deep-learning approach is its real-time operation capability. On the other hand, The CMCDOT framework does not identify the object in the urban scene. However, The CMCDOT represents the environment around the Ego-car through an occupancy grid map.  This map is a grid of cells, where each cell can identify occupancy states(i.e., static-object, dynamic-object, empty, and unknown) over time. Specifically, the static-object and dynamic-object states let us know more characteristics about the scene's objects, such as their position, motion estimation, etc. Finally, the achievement of the occupancy grid map is also in real-time based on the lidar data.

\footnotetext{The velocity and heading info are only for dynamic objects.}

\subsection{Fusion model}

\begin{figure}[h!]
\centering
\framebox{\includegraphics[height=6cm, width=8.5cm]{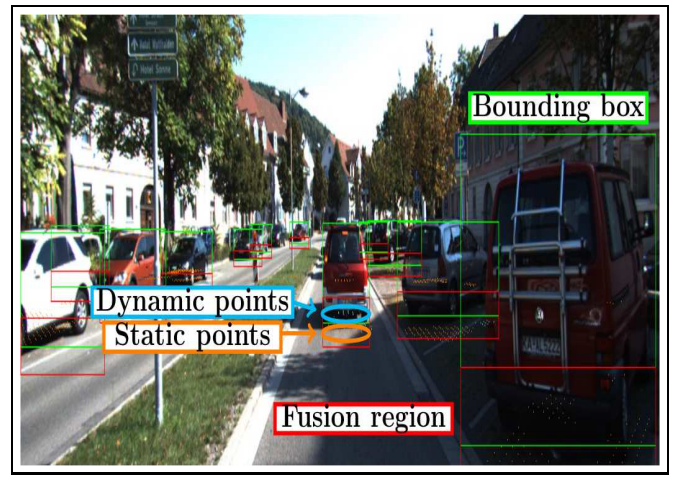}}
\caption{Urban scene representation considering the outcomes from YoloV3 and CMCDOT approaches. The \textit{green rectangles} represent the \textit{bounding boxes} generated by YoloV3. The \textit{Dynamic/Static points} are the information produced by the CMCDOT approach after the projective transformation. Finally, the \textit{red rectangles} represent the \textit{fusion regions}.}
\label{fig:FusionMo}
\end{figure}

In this part, we merged the outcomes from YoloV3 and the CMCDOT approaches to have a single representation from the environment. However, the result achieved from these two approaches is not in the same frame. Therefore, we had to align the results applying projective transformations. Our proposal projected the occupancy grid map accomplished by CMCDOT onto the image processed by the YoloV3 approach. In \cite{Gomez2020}, over section III-B, it is possible to find all the mathematical argumentation used to develop the projective transformations. Figure \ref{fig:FusionMo} shows the result obtained after aligning the outcome from YoloV3 and CMCDOT.

\begin{figure}[h!]
\centering
\framebox{\includegraphics[height=6cm, width=8.5cm]{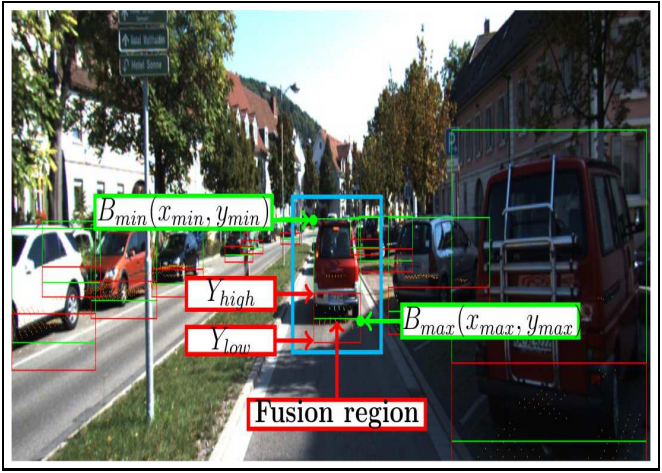}}
\caption{Definition of a fusion region regarding the bounding boxes obtained. The \textit{cyan rectangle} highlights a vehicle detected. The \textit{green points} represent the main bounding box points, ${ B }_{ min }$ and ${ B }_{ max }$, got from YoloV3. ${ Y }_{ high }$ and ${ Y }_{ low }$(i.e., defined in Equation \ref{eq:box}) are the high and low vertical thresholds used to determine the \textit{red rectangles} representing the fusion regions.}
\label{fig:Fumodel}
\end{figure}

We developed a fusion model that defines several regions of interest in the image to look for dynamic/static points inside them. We denominated those regions as \textit{fusion regions} (see Fig. \ref{fig:Fumodel}). Equation \ref{eq:box} describes the fusion model proposed.
\begin{equation}
{ \quad f }_{ m }\left( { B }_{ min,max }, { p }_{ i } \right) = \begin{cases} 1, { x }_{ min } < { x }_{ i } < { x }_{ max }  \\ \qquad and \quad { Y }_{ high } < { y }_{ i } < { Y }_{ low }. \\ 0, other \quad case. \end{cases}\\ 
\label{eq:box}
\end{equation}

where,

\begin{equation*}
    \begin{matrix} { Y }_{ high }\quad =\quad \frac { ((5*{ y }_{ max })-{ y }_{ min }) }{ 4 }  \\\\ 
    { Y }_{ low }\quad =\quad \frac { ((3*{ y }_{ max })+{ y }_{ min })  }{ 4 } \end{matrix}
\end{equation*}

According to equation \ref{eq:box}, the fusion model ${f}_{m}()$ depends from the bounding boxes points(see Fig. \ref{fig:Fumodel}) ${B}_{min}({x}_{min}, {y}_{min})$, ${B}_{max}({x}_{min},{y}_{min})$, and the dynamic/static points(see Fig. \ref{fig:FusionMo}) ${p}_{i}({x}_{i},{y}_{i})$. Furthermore, the ${f}_{m}()$ delimit the fusion regions(see Fig. \ref{fig:Fumodel}) among ${x}_{min}$, ${x}_{max}$, ${Y}_{high}$, and ${Y}_{low}$. When ${f}_{m}()$ found some points ${p}_{i}$ inside the fusion region, it classifies that point-set with the same label of the object detected by YoloV3. Otherwise, ${f}_{m}()$ discards the fusion regions where it did not find some point ${p}_{i}$. 

\subsection{Object information}

The outcome achieved in the fusion model part only classifies a point-set projected from an occupancy grid map. However, this point-set maintains a direct correspondence with the occupancy grid map's information. Therefore, it is possible to obtain the motion estimation, position 2D, velocity 2D, and heading from the point set classified. We called this procedure Point-wise features processing (PFP). Equation \ref{eq:points} describes the information related by each point belonging to the point-set classified. 

\begin{equation}
p_{i}^{c} = \left \{ x_{i},\quad y_{i},\quad x_{i}^{o},\quad y_{i}^{o},\quad vx_{i}^{o},\quad vy_{i}^{o} \right \}
\label{eq:points}
\end{equation}

In equation \ref{eq:points}, the character ${p}_{i}^{c}$ defines one point $i$ belonging to the point-set classified $c$. This point contains the coordinates ${x}_{i}$ and ${y}_{i}$ used in ${f}_{m}()$. Subsequently, the symbols ${x}_{i}^{o}$ and ${y}_{i}^{o}$ define the position 2D of the point ${p}_{i}^{c}$ over the occupancy grid map $o$. Finally, The denotations ${vx}_{i}^{o}$ and ${vy}_{i}^{o}$ represent the velocity 2D of ${p}_{i}^{c}$ over $o$. Those parameters also let us find the dynamic object's heading in the scene. ${vx}_{i}^{o}$ and ${vy}_{i}^{o}$ are equal zero when the point-set belong to a static-object.  

Considering all the features contained in the points ${p}_{i}^{c}$ classifies by ${f}_{m}$, the PFP begins to look for more details of the objects detected. For instance, the PFP defines the object motion estimation considering the cardinal number from the dynamic/static points-sets found in the fusion region(see Fig. 2). The higher cardinal value will determine the motion feature. Eventually, to find the object position, the PFP consider all the values ${p}_{i}^{c}({x}_{i}^{o},{y}_{i}^{o})$ from the point-set to compute their median value ${p}_{m}^{c}({x}_{m}^{o},{y}_{m}^{o})$. Similarly, to find the dynamic object velocity, the PFP also contemplates all the values ${p}_{i}^{c}({vx}_{i}^{o},{vy}_{i}^{o})$ from the point-set to obtain a median value ${p}_{m}^{c}({vx}_{m}^{o},{vy}_{m}^{o})$. Finally, for a dynamic object, the PFP computes the object's heading based on the arc tangent of ${vy}_{m}^{o}/{vx}_{m}^{o}$.

\subsection{Result}

\begin{figure}[h!]
\centering
\framebox{\includegraphics[height=6cm, width=8.5cm]{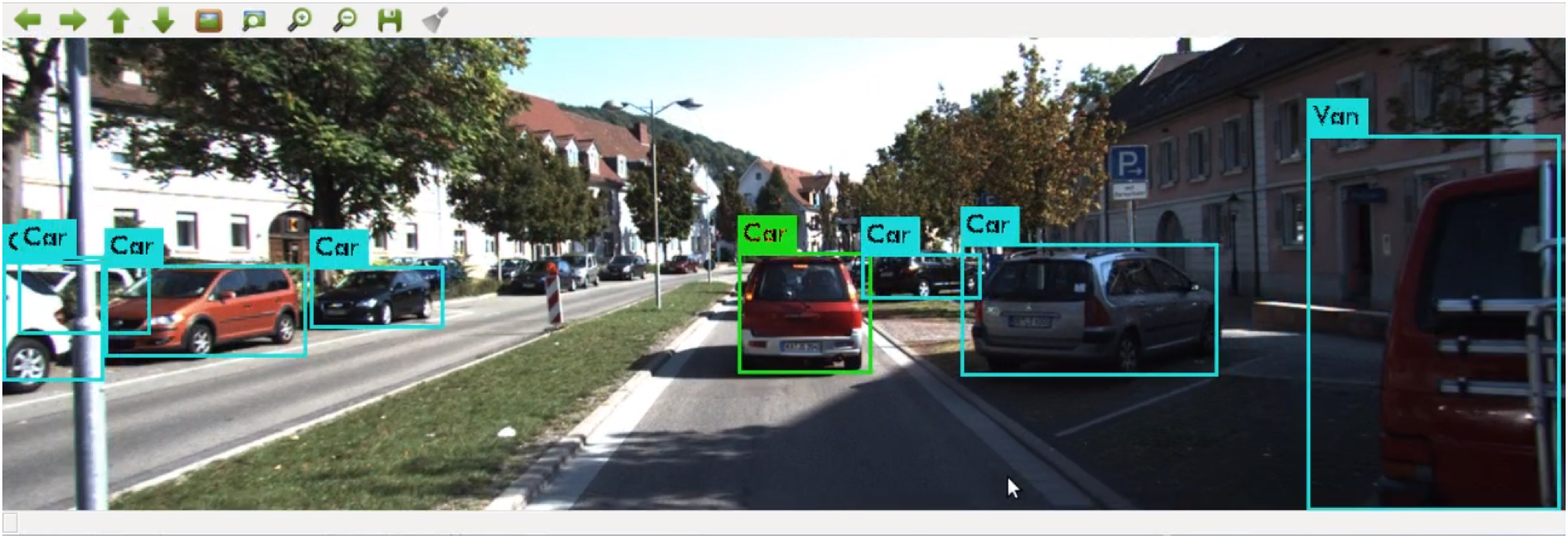}}
\caption{Graphical user interface used to present the result achieved with our proposal. The cars detected inside the \textit{blue bounding boxes} are static objects in the urban scene. On the other hand, the only dynamic object in the environment is the vehicle detected by the \textit{green bounding box}.}
\label{fig:result}
\end{figure}

We show the results obtained from the object information part through a Graphical User Interface (GUI) and ROS messages. Figure \ref{fig:result} shows some vehicles detected by our proposal in the GUI. While the GUI shows the result, Our algorithm also publishes ROS messages describing the motion estimation plus the object classification(e.g., Dynamic car), position 2D, velocity 2D, and heading. The results published in ROS corresponding to each object detected in front of the Ego-car over time.  

It is essential to highlight that we didn't consider the object's position, velocity, and heading in the experiment results. The main reason is that the dataset used as a benchmark doesn't provide this type of information about the objects.


\section{EXPERIMENTS}

This section describes the datasets used, the experimental setup contemplated, and the results obtained with our object detection approach. 
\subsection{Datasets}

We evaluated our approach using the \textit{KITTI} \cite{Geiger2013} and the \textit{KITTI MOD} \cite{siam2018} datasets. We used the training and test images from KITTI to train the YoloV3 approach. Moreover, in our proposal's evaluation, we employed lidar data from one city category. Concerning the KITTI MOD dataset, we worked with it because it was necessary to recognize the objects' motion based on the KITTI dataset. The KITTI MOD dataset has 5997 static vehicles and 2383 dynamic ones labeled.   
\subsection{Experimental Setup}

\begin{figure}[h!]
\centering
\framebox{\includegraphics[height=6cm, width=8.5cm]{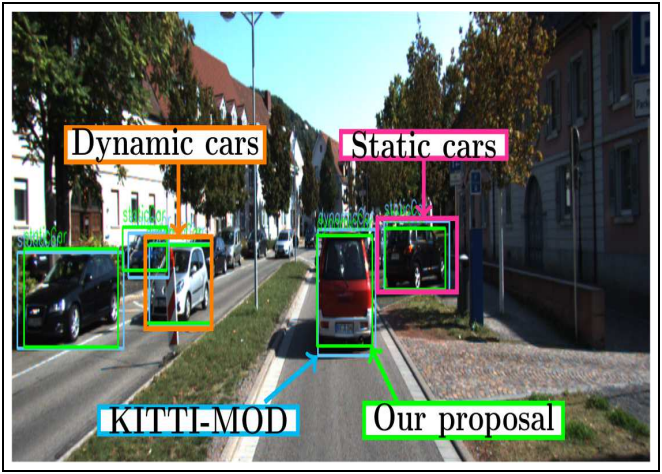}}
\caption{Example of the experiment accomplished. The \textit{orange rectangle} contains one of the dynamic vehicles in the urban scene. Otherwise, the \textit{magenta rectangle} represents the static vehicles. \textit{Blue and green rectangles} in the image illustrate the comparison between the objects labeled from the KITTI MOD dataset and the objects detected by our proposal.}
\label{fig:expKM}
\end{figure}

To Evaluate the object detection and the motion estimation of our approach, we used ROS to publish several urban images from KITTI MOD with their corresponding lidar data from KITTI. We obtain this ordered image sequence and their lidar data in our algorithm through ROS messages. Inside that urban environment, all the objects are dynamic or static vehicles. This fact is because the KITTI MOD dataset recognizes only vehicles' motion. Figure \ref{fig:expKM} shows an example of our test. 

In Fig \ref{fig:expKM}, it is possible to observe some dynamic and static vehicles around the Ego-car. Furthermore, the figure also shows the application of the mean average precision (mAP) evaluation in the experiment(i.e., blue and green labels in Fig \ref{fig:exp}). For the mAP evaluation, we considered the open-source tool developed in \cite{Gomez2020}. In the mAP evaluation, we examined all the vehicles inside a longitudinal distance of thirty meters from the Ego-car. The main reason is that we found in \cite{Cartucho2018} that the YoloV3 approach has a better performance in that distance. In \cite{rangesh2019}, also is contemplated a similar longitudinal distance for the object detections.

\subsection{Experimental results}

\begin{figure}[H]
\centering
\framebox{\includegraphics[height=6cm, width=8.5cm]{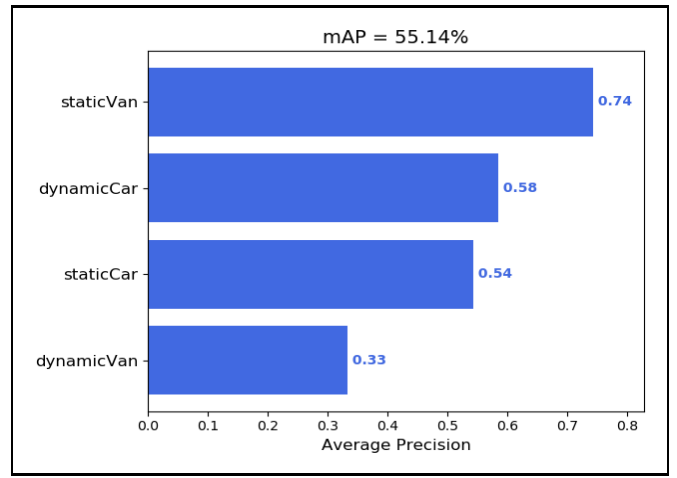}}
\caption{Mean average precision (mAP) result achieved in our test. The figure shows the Average Precision (AP) values(i.e., in \textit{blue bars}) obtained by each object detected during the experiment, considering their class and motion estimation.}
\label{fig:exp}
\end{figure}

The result obtained based on our experiment setup are present in detail through Tables \ref{tab:metrics}-\ref{tab:modnet} and Figure \ref{fig:exp}. Table \ref{tab:metrics} show the performance evaluation results based on the precision, recall, and F1 score metrics. The four classes obtained during the experiment also include their motion estimation. Table \ref{tab:metrics} also reveals that the F1 score metric maintains a balance between the precision and recall values obtained in the three first classes. The class \textit{dynamicCar} is the most significant example with an F1-score value of 73\%.

\begin{table}[h]
\centering
\caption{Performance result of our approach.}
\label{tab:metrics}
\begin{tabular}{|c|c|c|c|}
\hline
\textbf{Class}      & \textbf{Presicion} & \textbf{Recall} & \textbf{F1-score} \\ \hline
\textbf{staticVan}  & 83                 & 75              & 79               \\ \hline
\textbf{dynamicCar} & 72                 & 74              & 73               \\ \hline
\textbf{staticCar}  & 91                 & 58              & 71               \\ \hline
\textbf{dynamicVan} & 100                & 33              & 50               \\ \hline
\end{tabular}
\end{table}

Figure \ref{fig:exp} shows the performance evaluation of our proposal using the mean average precision (mAP). The graph also reports average precision (AP) achieved by each object in the test based on their class and motion estimation. It is possible to observe that the AP value between the classes \textit{dynamicCar} and \textit{staticCar} are near. On the other hand, in Table \ref{tab:modnet} we compare the mAP obtained by our approach against the MODNET \cite{siam2018} approach. Although we achieved an mAP nearby, our proposal also can obtain more characteristics from the object detected(i.e., position, velocity, and heading). It is possible to find all the information about the experiment results in \url{https://gitlab.inria.fr/agomezhe/paper_results}. 

\begin{table}[h!]
\centering
\caption{Quantitative evaluation on KITTI MOD data of our proposal.}
\label{tab:modnet}
\begin{tabular}{|c|c|}
\hline
\textbf{Method}                      & \textbf{mAP}   \\ \hline
MODNET\cite{siam2018}                & \textbf{62.57} \\ \hline
Ours                                 & 55.14          \\ \hline
\end{tabular}
\end{table}

\begin{figure}[h]
\centering
\framebox{\includegraphics[height=6cm, width=8.5cm]{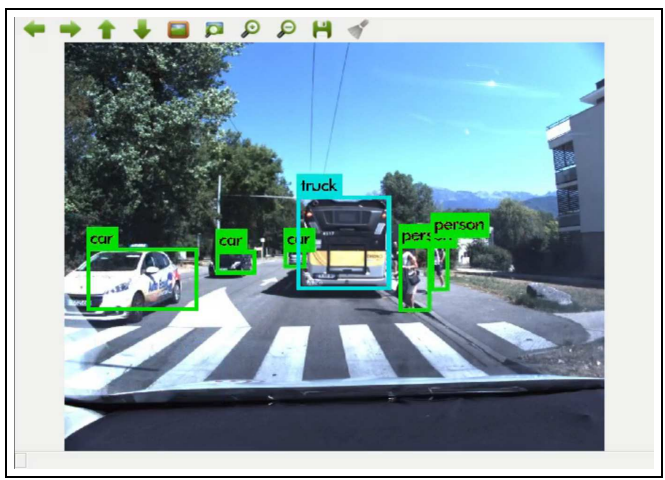}}
\caption{Illustration of our proposal using the autonomous vehicle platform. The graph shows several \textit{green bounding boxes} representing different dynamic objects around our autonomous vehicle platform. The \textit{blue bounding box} present in the image exhibit only one static object.}
\label{fig:Fusion}
\end{figure}

Finally, Figure \ref{fig:Fusion} shows a demonstration example of our proposal with results achieved.  In this urban demonstration, we used sensor data from our autonomous vehicle platform. The main objective was to test our approach considering different urban situations under extreme weather conditions(i.e., sunny and rainy weather). It is possible to find a video with the demonstration results in \url{https://youtu.be/Rd-0B0--mlc}.

\section{CONCLUSIONS}

This paper considers detecting dynamic and static objects around an autonomous vehicle in an urban environment. However, our proposal is not only limited to object detection and its motion estimation. Furthermore, we also look for the position, velocity, and heading from the objects detected. We found all these characteristics of the objects detected in the scene, considering the fusion of the YoloV3 and CMCDOT approaches' outcomes. That information lets to define the fusion regions and the point-wise features utilized in our approach.

The results obtained show that our proposal is nearby to the performance achieved by the work used as a reference. Therefore, to improve our approach's performance, future works are focus on taking some actions like updating the YoloV3 to YoloV4 object detector. Moreover, it will also be necessary to train the YoloV4 with a new balanced dataset. Finally, even when all our approach was optimized using a GPU, we will need to compensate our hardware disadvantage compared to the referenced work. 

\addtolength{\textheight}{-12cm}   


\section*{ACKNOWLEDGMENT}

This work has been supported by the French Government in the scope of the FUI STAR project. We also would like to thank \"{O}zg\"{u}r Erkent for his meaningful technical discussions and suggestions.


\end{document}